\documentclass{sigchi-ext}
\usepackage[T1]{fontenc}
\usepackage{textcomp}
\usepackage[scaled=.92]{helvet} 
\usepackage{graphicx} 
\usepackage{balance}  
\usepackage{booktabs} 
\usepackage{ccicons}  
\usepackage{ragged2e} 

\def\plaintitle{SIGCHI Extended Abstracts Sample File: Note Initial
  Caps} 
\def\emptyauthor{}
\def\plainkeywords{generative AI; large language models; role-playing; transparency; explainability; social attribution; artificial intelligence}

\title{
Addressing Social Misattributions of Large Language Models: An HCXAI-based Approach}

\numberofauthors{3}
\author{%
  \alignauthor{%
    \textbf{Andrea Ferrario}\\
    \affaddr{Institute of Biomedical Ethics \ and History of Medicine, University of Zurich and ETHZ,} \\
    \affaddr{Zurich, Switzerland.} \\
    \email{andrea.ferrario@ibme.uzh.ch} }\alignauthor{%
    \textbf{Alberto Termine}\\
    \affaddr{Dalle Molle Institute for Artificial Intelligence (IDSIA USI-SUPSI),}\\
    \affaddr{Lugano, Switzerland, and}\\
    \affaddr{College of Humanities, Ecole Polytechnique Federale de Lausanne (EPFL),}\\
    \affaddr{Lausanne, Switzerland.}\\
    \email{alberto.termine@idsia.ch} } \vfil \alignauthor{%
    \textbf{Alessandro Facchini}\\
   \affaddr{Dalle Molle Institute for Artificial Intelligence (IDSIA USI-SUPSI),}\\
    \affaddr{Lugano, Switzerland.}\\
    \email{alessandro.facchini@idsia.ch}} \vfil \alignauthor{%
    } }

\definecolor{linkColor}{RGB}{6,125,233}
\usepackage{hyperref}
\hypersetup{%
  pdftitle={\plaintitle},
  pdfauthor={\emptyauthor},
  pdfkeywords={\plainkeywords},
  bookmarksnumbered,
  pdfstartview={FitH},
  colorlinks,
  citecolor=black,
  filecolor=black,
  linkcolor=black,
  urlcolor=linkColor,
  breaklinks=true,
}

\begin{document}

\CopyrightYear{2024}
\setcopyright{rightsretained}
\conferenceinfo{ACM CHI Workshop on Human-Centered Explainable AI (HCXAI24)}{}
\isbn{978-1-4503-6819-3/20/04}
\doi{https://doi.org/10.1145/3334480.XXXXXXX}
\copyrightinfo{\acmcopyright}

\maketitle

\RaggedRight{}

\begin{abstract}
Human-centered explainable AI (HCXAI) advocates for the integration of social aspects into AI explanations. Central to the HCXAI discourse is the Social Transparency (ST) framework, which aims to make the socio-organizational context of AI systems  accessible to their users. In this work, we suggest extending the ST framework to address the risks of social misattributions in Large Language Models (LLMs), particularly in sensitive areas like mental health. In fact LLMs, which are remarkably capable of simulating roles and personas, may lead to mismatches between designers' intentions and users' perceptions of social attributes, risking to promote emotional manipulation and dangerous behaviors, cases of epistemic injustice, and unwarranted trust. To address these issues, we propose enhancing the ST framework with a fifth 'W-question' to clarify the specific social attributions assigned to LLMs by its designers and users. This addition aims to bridge the gap between LLM capabilities and user perceptions, promoting the ethically responsible development and use of LLM-based technology.
\end{abstract}




\newpage
\section{Introduction}
Research has recently started investigating artificial intelligence (AI) under a socio-technical lens, attempting to contextualize this technology within its broader social and organizational environment. From the `fruitful collaboration' between sociology and computer science \cite{sartori2022sociotechnical}, the perspective that AI systems are artefacts embedded in a network of norms that shape their design and influence trust in them has made its way in the scientific discourse \cite{Jacovi2021,ehsan2021expanding,benk2022value}. In particular, human-centered explainable artificial intelligence (HCXAI), which is promoted by initiatives, such as the ACM CHI Workshop on Human-Centered Explainable AI,\footnote{\url{https://hcxai.jimdosite.com/}} focuses on the necessity to consider the social component of \emph{explaining} how AI works. Prominently, Ehsan et al.'s Social Transparency framework  integrates socio-organizational contexts into AI-mediated decision-making, aiming to make the technological, decision-making, and organizational contexts visible and understandable \cite{ehsan2021expanding}. 

In these notes, we argue that Social Transparency can be used to address the risks stemming from \textbf{social misattributions} of Large Language Models (LLMs). Our argument goes as follows. First, following Shananan et al.'s work \cite{shanahan2023role}, we note that LLMs are essentially role-play devices, to which we can assign (1) roles and (2) personas. Roles are the expected behaviors of the LLMs within the socio-technical context they operate in. With the term `persona' we denote the \textit{social face} \cite{Goffman55,goffman1967}-- e.g., personality traits, such as being curious, polite and empathetic--that LLMs are required to express while simulating individuals in conversations with their users.\footnote{Due to lack of space, we refer to Bargiela Chiappini and Haugh's work \cite{bargiela2009face} for a detailed discussion on the concept of face.} 
LLMs can follow role and persona assignments--in short: \emph{social attributions}--thanks to their notable capability to perform a variety of downstream tasks in different contexts \cite{balas2023exploring,ferrario2024large,chen2023llm}. In particular, this  capability affords humans a certain degree of flexibility in performing social attributions of LLMs. This is a source of notable risks. 
In fact, on the one hand, designers assign intended roles and personas to LLM-based applications, such as an `empathetic psychiatrist' \cite{chen2023llm}. On the other hand, users may perform different attributions depending on their perception of the abilities of the LLMs. However, these attributions may descend from incorrect perceptions of the \textit{objective} capabilities of an LLM. This problem, we argue, is not restricted to the case of humans interacting with LLM-based applications. However, the unprecedented ability of LLMs to simulate roles and human personality traits, including being an epistemic authority in a domain and expressing `empathy' \cite{chen2023llm,ferrario2023experts}, and the potential diffusion of these systems to ethically-sensitive domains, such as mental health and medical ethics \cite{balas2023exploring,ferrario2024large}, calls for timely addressing the risk of social misattributions of LLMs.\footnote{For instance, the risk of social misattributions of AI systems predicting small bone fractures from medical images or inappropriate comments on social media platforms is arguably smaller than in the case of LLM-based applications. Similarly, we believe that `traditional' conversational artificial intelligence agents cannot reach the level of stylistic flexibility and conversational prowess shown by LLM-based systems (especially after their fine-tuning).}
These risks include a higher propensity of being nudged, the reinforcement of negative behaviors \cite{weidinger2022taxonomy}, cases of epistemic injustice \cite{laacke2023bias}, wrong accountability attributions \cite{ferrario2023experts}, and, in general, unwarranted trust in LLM-based applications \cite{Jacovi2021,ferrario2022explainability}. Hence,
developing effective strategies to counter social misattributions of LLMs is key to develop LLM-based applications responsibly. 

Finally, to address social misattributions of LLMs we suggest to extend the Social Transparency framework by including a fifth `W-question' to its `4W model' \cite{ehsan2021expanding} thus clarifying to the users of an LLM-based application \emph{which} social attribution
is actually assigned to the model and \emph{which} ones are promoted by its users instead. We elaborate on our proposal by introducing two methods to support the provision of information that allows answering the fifth `W-question' in real-world applications.

\section{LLMs, functions and role-playing}\label{sec:Roles}
LLMs are a type of generative AI performing context-aware text generation \cite{weidinger2022taxonomy}. Authors state that these models answers user queries similarly to an autocomplete function of highly sophisticated search engines \cite{floridi2023ai}. However, this perspective on the function of LLMs is somehow limiting. In fact, LLMs serve \textit{different} functions. To elaborate on this point, following Crilly \cite{crilly2010roles}, we promote a separation between technical and non-technical functions of AI technology. First, the \textbf{techno-function} of an LLM is to compose textual outputs by computing the empirical probability distribution of the `next token' after being trained on massive text corpora \cite{shanahan2023role}. This function is  objective, insofar as it is independent on how different users interact with the LLM in different contexts. In addition to their techno-function, however, LLMs perform \textbf{socio-functions} \cite{crilly2010roles}, namely, functions that hinge on the social capabilities of these  systems, which depends on both its technical function and the social norms that shape the use of an LLM in a given context. One prominent example of LLM socio-function is `role-playing', that is the function of simulating different roles and personas in a conversational setting \cite{shanahan2023role}. In fact, by means on an attribution of a role and a persona--in short: a \textit{social attribution}--users may engage themselves in context-aware and somehow realistic conversations with these systems.

\section{Social misattributions of LLMs}
Social attributions have a normative component: an LLM endowed with a given role and persona is expected to perform as such. On the one hand, an `empathetic psychoterapist' LLM will need to simulate active listening and
express concern towards the needs of its users. On the other hand, simulating an `inspired Renaissance madrigalist' will require the generation of compelling examples of madrigal poetry. Social attributions are subjective processes, shaped by social and cultural factors that may change over time. To this end, the emersion of roles and personas  needs consolidation, approval, and endorsement. This takes into account pragmatic considerations, including the utility of a role to designers and users. In general, the rules and laws of organizations and societies determine the emersion of socially-accepted roles and personas for technology, including LLM-based applications.\footnote{This is a well-known trope in the science fiction literature. Consider, for instance, the repulsion towards artificial intelligence in F. Herbert's Dune series \cite{herbert1965dune} and the hatred towards A.I.--here, `Abominable Intelligence,'--manifested by the `Imperium of Man' in the Warhammer 40,000\texttrademark~ universe.} 

While research is starting exploring which roles and personas should be assigned to LLM-based applications \cite{levkovich2023identifying,chen2023llm,balas2023exploring,ferrario2024large,ferrario2023experts}, here, we highlight two problems emerging from social attributions. First,  designers and users may promote different social attributions for the same LLM-based application.
Second, users' social attributions may descend from incorrect perceptions of the model's capabilities. As a result, users can hold unwarranted expectations about the behaviour of these applications and be affected by detrimental consequences. Below, we elaborate on these points in an ethically challenging scenario.

\subsection{An example from the mental health domain}
Consider the case of an individual with depression who attributes the role of `digital psychiatrist' to an LLM-based application, which, in turn, is meant by its designers to provide therapeutic support only. The role attribution is inappropriate for different reasons. First, LLMs cannot be epistemic experts in the domain of psychiatry \cite{ferrario2023experts}. They do not possess genuine understanding of their users' queries as, essentially, they autocomplete their responses based on stochastic computations and lack key abilities that characterize human experts, such as conscientiousness, intellectual curiosity and perseverance \cite{ferrario2023experts}. In addition, differently from psychiatrists, AI systems cannot prescribe medications and do not follow a code of conduct.\footnote{Simply encoding  a few rules in an LLM, such as `do not prescribe medications' or `always maintain patient  confidentiality' does not exemplify following psychiatrists' role of conduct.} In summary, LLM-based applications lack key attributes that constitute the \textit{role of a psychiatrist}. Further, let us suppose that the user assigns an empathetic, caring and accepting persona to the LLM-based application after a few felicitous interactions with the system. Despite this attribution, \textit{none of these personality traits can be genuinely simulated by an LLM}. In fact, they require an understanding of the users' psychology, social context and experiences that lies beyond the capabilities of the techno-function of these models. This said, we contend that LLMs can be sometimes attributed with certain personas. In fact, examples of personality traits that can be simulated by LLMs include being concise, verbose or accurate in their responses. Simply, those assigned by the user in our example constitute a case of social misattribution.
As a result, the user expects the LLM-based application to suggest taking actions on the basis of an in-depth expertise in the domain of psychiatry (role), a genuine understanding of the depressed individual's emotional states and a willingness to help them that is independent of their personal history and experiences (persona). This is neither what the LLM can technically perform nor the intended social attribution that designers implemented during the design of the system. 

The risks posed by social misattributions of LLMs can be significant. For instance, research has shown that ChatGPT-3.5 prescribed medications to individuals affected by anxiety or depression, despite this being not allowed to such systems \cite{farhat2023chatgpt}. In addition, believing that LLMs are empathetic and caring professionals exposes vulnerable individuals to being nudged and emotionally manipulated. As a result, the provision of inappropriate responses, incorrect information or dangerous recommendations by these systems can lead to substantive harm.
In general, social misattributions of LLMs lead to unwarranted trust in these systems \cite{Jacovi2021,ferrario2022explainability}. Here, 
a trusting relation between a user of an LLM-based application and the system is unwarranted if it not grounded in objective capabilities, e.g, being reliable, accurate or providing information that supports transparency, that the system is supposed to maintain during the interactions \cite{Jacovi2021,ferrario2022explainability}. Indeed, the attribution of a role and persona generates the expectation in users that an LLM will behave as to fulfill those social positions, i.e., that it will be able to perform the tasks that an agent playing that role is supposed to perform showing the traits associated with that persona.
However, if the LLM does not really possess the capabilities necessary to fulfill the attributed role  and persona 
and just seems to simulate some of their traits to a certain degree, users' expectations will be disappointed and their trust unwarranted. Then, if fostering warranted trust in AI systems is a desideratum of human-AI interactions \cite{Jacovi2021} and a necessary condition for the responsible use of AI in society, it is imperative to address the problem of social misattributions of LLMs.

\section{Adapting Social Transparency to address social misattributions: the 5W model }\label{sec:SocialTransparencyPractices}
Ehsan et al.'s Social Transparency framework hinges on a model of explanations of the outputs computed by AI systems that is called the `4W model' \cite{ehsan2021expanding}. The 4W model pertains four `W-questions', aiming to explain `\emph{who} did \emph{what} with the AI system, \emph{when}, and \emph{why} they did what they did— in order to have adequate socio-organizational context around
the AI-mediated decisions' \cite[pag. 5, emphasis in original]{ehsan2021expanding}. 
To address the problem of social misattribution of LLMs, our proposal is to adapt the Social Transparency framework by augmenting the `4W' model to a `5W' model. The additional `W-question' focuses on identifying (1) \emph{which} social attributions are justified for a LLM in a certain context, and (2) \emph{which} social attributions a user assigns to the LLM in the same context.

To support the integration of the which' question into our 5W model, we sketch two methodologies whose detailed analysis and development we reserve for future work.

\textbf{Developing a taxonomy of social attributions.} 
Organizations that develop LLM-based applications or use them for their services and products should provide taxonomies of the \emph{appropriate} and \emph{inappropriate} roles and personas for their systems.\footnote{The term ``appropriate'' refers to normative conditions
that take into account the limits of the capabilities of LLMs in their context of applicability.}
In particular, these taxonomies should include 
examples of (in-)appropriate social attributions to guide users through the process of role and persona assignment.
For instance, users could benefit from understanding that the lack of effective measures against adversarial attacks in an LLM-based application makes the attribution of a `polite' persona to the system untenable over time. The development of social attribution taxonomies should take place in the context of a general promotion of the social transparency of AI systems, which can be carried out according to the methodology of \emph{participatory design} \cite{robertson2012participatory}. Key here is providing  descriptions of social attributions, which take into account the different users' socio-cultural perspectives on the possible roles and personas of the LLM-based application.  Thus, the development of an LLM-based application should involve a trans-disciplinary team, including experts in the epistemology, psychology and sociology of AI, as well as human-computer interaction. 

\textbf{Implementing techniques that detect and prevent social mis-attributions.} Further, by developing algorithms that detect potential misattribution \emph{dynamically} in the conversations that users have with LLM-based applications \cite{Zhang2023}, mitigation strategies against social misattributions could be implemented. For instance, 
the LLM-based application could simply warn users that they are likely incurring in an inappropriate social attribution of the system if a `risk threshold' is met and refer to the taxonomy of social attributions to provide more information on this challenge. Finally, the efficacy of this approach at reducing cases of social misattributions can be tested against more traditional, static strategies, i.e., the use of disclaimers on appropriate roles and personas on the application interface or at the beginning of each conversation with users.\footnote{ChatGPT3.5/4 implement a similar, static strategy to warn users about the possibility to provide erroneous information in their responses by stating that  ``ChatGPT can make mistakes. Consider checking important information.''}

\balance{} 

\bibliographystyle{SIGCHI-Reference-Format}
\bibliography{biblio}

\end{document}